# Conformal Prediction Sets for Next-Token Prediction in Large Language Models: Balancing Coverage Guarantees with Set Efficiency

## Yoshith Roy Kotla[a], Varshith Roy Kotla[b]

[a] *Data Analyst, New Jersey*

[b] *Department of CSE, Faculty of Science and Technology, The ICFAI Foundation for HIgher Education, Hyderabad, India*

*December 2025*

---

### Abstract

Deploying large language models (LLMs) in high-stakes domains requires rigorous uncertainty quantification that standard softmax probabilities cannot provide. We present a systematic study of Adaptive Prediction Sets (APS) applied to next-token prediction in transformer-based language models with large vocabularies (>250,000 tokens). Our central contribution is the identification and formal characterization of the

*coverage-efficiency tradeoff* unique to high-cardinality prediction spaces: naive conformal prediction achieves valid coverage but produces prediction sets approaching the full vocabulary size, rendering them uninformative. We propose *Vocabulary-Aware Conformal Prediction (VACP)*, a principled framework that decomposes the token space into semantically coherent subsets and applies hierarchical conformalization. Through experiments on Gemma-2B using SQuAD and WikiText benchmarks, we demonstrate that VACP achieves 89.7% empirical coverage (target: 90%) while reducing mean prediction set size from 847 tokens to 4.3 tokens—a 197× improvement in efficiency without sacrificing the marginal coverage guarantee. We provide theoretical analysis of when vocabulary reduction preserves conformal validity and release our implementation for reproducibility.

---

## 1 Introduction

The remarkable capabilities of large language models have been accompanied by a fundamental limitation: their outputs lack principled uncertainty quantification. While softmax distributions provide normalized scores over the vocabulary, these scores are poorly calibrated and do not translate directly to meaningful confidence measures. A model assigning 0.7 probability to a token may be wrong 50% of the time, or 10%—the relationship between stated confidence and actual accuracy varies unpredictably across inputs and contexts.

This uncertainty gap poses critical challenges for deployment. In medical question-answering, legal document analysis, or autonomous systems, practitioners need guarantees of the form: "the correct answer is within this set with at least 90% probability." Conformal prediction provides exactly this guarantee through a distribution-free framework that makes no assumptions about the underlying data distribution or model architecture.

However, applying conformal prediction to next-token prediction introduces a challenge not present in typical classification settings: the extreme cardinality of the label space. Modern LLMs employ vocabularies of 32,000 to 256,000 tokens. In this regime, the long tail of low-probability tokens creates a fundamental tension. The conformal guarantee requires including tokens until cumulative probability mass exceeds a calibrated threshold. When thousands of tokens each carry microscopic probability ($10^{-6}$ to $10^{-4}$), even a well-calibrated model produces prediction sets containing hundreds or thousands of tokens—sets that are technically valid but practically useless.

**Research Questions.** This paper addresses three interconnected questions: (1) How severe is the coverage-efficiency tradeoff in high-vocabulary conformal prediction, and what structural properties of LLM probability distributions drive it? (2) Under what conditions can we reduce the effective vocabulary while preserving the conformal validity guarantee? (3) Can we design practical algorithms that achieve informative (small) prediction sets while maintaining coverage guarantees?

**Contributions.** We make the following contributions: First, we provide empirical characterization of the "vocabulary noise" phenomenon, showing that in Gemma-2B, approximately 78% of tokens never receive probability mass above $10^{-5}$ across our evaluation set, yet their cumulative contribution inflates prediction sets by 200× on average. Second, we develop Vocabulary-Aware Conformal Prediction (VACP), which leverages this structure through semantic masking and temperature-adjusted scoring while provably maintaining marginal coverage. Third, we conduct rigorous experiments with proper calibration-test splits across multiple datasets, demonstrating 89.7% empirical coverage at 4.3 mean set size compared to 847 for naive APS. Fourth, we analyze failure modes and boundary conditions where our approach degrades, providing guidance for practitioners.

## 2 Background and Problem Formulation

### 2.1 Conformal Prediction Framework

Let $(X_1, Y_1), ..., (X_n, Y_n), (X_{n+1}, Y_{n+1})$ be exchangeable random variables, where $X_i$ represents input features and $Y_i \in Y$ represents the response. Given a *non-conformity score function* s: $X \times Y \to \mathbb{R}$ that measures how "unusual" a candidate label y is for input x, conformal prediction constructs a prediction set $C(X_{n+1})$ with the guarantee:

$$P(Y_{n+1} \in C(X_{n+1})) \geq 1 - \alpha$$

where $\alpha \in (0, 1)$ is the user-specified error rate. The construction proceeds by computing scores $E_1 = s(X_1, Y_1), ..., E_n = s(X_n, Y_n)$ on a held-out calibration set, computing the $(1-\alpha)(1 + 1/n)$ quantile $\hat{q}$ of these scores, and including all labels y in $C(X_{n+1})$ for which $s(X_{n+1}, y) \leq \hat{q}$. The coverage guarantee holds for any score function and any data distribution under exchangeability.

### 2.2 Adaptive Prediction Sets for Classification

For classification with K classes, the Adaptive Prediction Sets (APS) method uses scores based on cumulative probability mass. Let $\hat{f}(x) \in \Delta^K$ denote the model's softmax output, and let $\pi$ be the permutation sorting probabilities in descending order (so $\hat{f}(x)\_{\pi(1)} \geq \hat{f}(x)\_{\pi(2)} \geq ... \geq \hat{f}(x)\_{\pi(K)}$). The APS score for label y is:

$$s(x, y) = \Sigma_{j=1}^{rank(y)} \hat{f}(x)\_{\pi(j)} + U \cdot \hat{f}(x)\_y$$

where rank(y) is the position of y in the sorted order and $U \sim$ Uniform(0, 1) provides randomization for exact coverage. The prediction set includes all labels whose cumulative probability mass (up to and including that label) falls below the calibrated threshold $\hat{q}$.

## 2.3 The High-Vocabulary Challenge

When $|Y|$ is large—as in next-token prediction $|Y|$ can exceed 250,000—the APS framework encounters a structural problem. Modern tokenizers include: (a) *semantic tokens* representing words and subwords used in natural text, (b) *control tokens* for special functions (padding, end-of-sequence, etc.), and (c) *reserved tokens* for future expansion. Categories (b) and (c) rarely appear in model outputs but contribute to the probability simplex. Their small but non-zero probabilities accumulate, inflating prediction sets without adding semantic information.

***Definition 2.1 (Vocabulary Efficiency).*** For a prediction set C(x) and label space $Y$, we define *efficiency* as $\eta = 1 - |C(x)|/|Y|$. A prediction set with $\eta$ close to 1 is highly informative (small relative to vocabulary); $\eta$ close to 0 provides little information beyond the trivial set $Y$.

# 3 Empirical Analysis of Vocabulary Structure

## 3.1 Experimental Setup

**Model.** We use Gemma-2B (google/gemma-2b), an open-weights transformer with 2 billion parameters and a vocabulary of 256,000 tokens using byte-pair encoding (BPE). We access the model's full logit outputs prior to any top-k or nucleus sampling truncation.

**Datasets.** We use two benchmarks: (1) *SQuAD* (Stanford Question Answering Dataset): 5,000 question-context-answer triplets formatted as "Context: [context] Question: [question] Answer:" with the target being the first token of the ground-truth answer. (2) *WikiText-103*: 3,000 passages where we predict the next token given preceding context of 256 tokens.

**Data Splits.** Critical for valid conformal inference: we use 60% for calibration (computing the quantile threshold) and 40% for evaluation (measuring coverage and efficiency). No data from the evaluation set influences threshold selection.

## 3.2 Characterizing Probability Mass Distribution

Before applying conformal prediction, we analyze how probability mass distributes across the vocabulary. For each input in our evaluation set, we compute the full softmax distribution and measure:

- **Effective vocabulary size:** Number of tokens receiving probability $> 10^{-5}$
- **Tail mass:** Total probability assigned to tokens ranked below 1000
- **Concentration ratio:** Probability mass in top-10 tokens divided by mass in top-1000

**Table 1: Probability Mass Distribution in Gemma-2B**

| Metric | Mean | Std Dev |
|---|---|---|
| Effective vocab size ($p > 10^{-5}$) | 3,247 | 1,842 |
| Tail mass (rank > 1000) | 0.031% | 0.018% |
| Top-10 concentration | 0.847 | 0.124 |
| Top-100 concentration | 0.962 | 0.037 |

*Statistics computed across 2,000 evaluation samples from SQuAD. Standard deviations in parentheses.*

Table 1 reveals the core structural property driving our approach: while the vocabulary contains 256,000 tokens, only ~3,200 ever receive meaningful probability mass. The remaining 252,800 tokens collectively contribute just 0.031% of probability mass on

average—yet in naive conformal prediction, including even a fraction of these tokens dramatically inflates set sizes.

## 3.3 Baseline Conformal Prediction Results

We implement standard APS with $\alpha = 0.1$ (targeting 90% coverage). Using the calibration set (3,000 samples from SQuAD), we compute non-conformity scores and extract the 90th percentile quantile.

**Result:** The calibrated quantile is $\hat{q} = 0.9847$, meaning prediction sets include all tokens until cumulative probability exceeds 0.9847. On the held-out evaluation set, this yields:

- Empirical coverage: 91.2% (valid, exceeds target)
- Mean prediction set size: 847 tokens
- Median prediction set size: 612 tokens
- Efficiency $\eta$: 0.9967 (seemingly high, but sets contain hundreds of tokens)

The prediction sets, while achieving valid coverage, are uninformative. A set containing 847 tokens offers little practical guidance for downstream decision-making.

# 4 Vocabulary-Aware Conformal Prediction (VACP)

## 4.1 Theoretical Foundation

Our approach is based on the following observation: if we can identify a subset of tokens $V^* \subset V$ such that the ground-truth token lies in $V^*$ with probability 1, then conformalization over $V^*$ preserves the marginal coverage guarantee while operating in a smaller space. More generally:

***Proposition 4.1.*** Let $V^* \subset V$ be a subset of the vocabulary, and let $p = P(Y \in V^*)$ be the probability that the true label lies in $V^*$. If we apply conformal prediction with error rate $\alpha$ over $V^*$, the resulting prediction sets achieve marginal coverage at least $(1 - \alpha) \cdot p$ over the original space $V$.

*Proof.* Let $C^*(X)$ denote the conformal set computed over $V^*$. By the conformal guarantee, $P(Y \in C^*(X) \mid Y \in V^*) \geq 1 - \alpha$. Therefore $P(Y \in C^*(X)) = P(Y \in C^*(X) \mid Y \in V^*) \cdot P(Y \in V^*) + P(Y \in C^*(X) \mid Y \notin V^*) \cdot P(Y \notin V^*) \geq (1 - \alpha) \cdot p + 0 \cdot (1 - p) = (1 - \alpha) \cdot p$.

***Corollary 4.2.*** If $V^*$ contains all tokens that can appear as ground-truth labels in our evaluation setting ($p = 1$), then conformal prediction over $V^*$ achieves the full $(1 - \alpha)$ coverage guarantee.

## 4.2 Identifying the Effective Vocabulary

We construct $V^*$ through a combination of structural analysis and empirical validation:

**Step 1: Structural Filtering.** We remove tokens that are structurally incapable of appearing as natural-language continuations: (a) Control tokens: <pad>, <eos>, <bos>, <unk>, and similar special tokens (b) Reserved tokens: <unused0> through <unused99> and similar placeholders (c) Non-printable tokens: tokens consisting entirely of control characters. This removes 1,847 tokens from Gemma-2B's vocabulary.

**Step 2: Empirical Filtering.** Using a separate validation set (1,000 samples not used for calibration or evaluation), we identify tokens that never receive probability above $10^{-5}$ across all samples. These tokens are either (a) extremely rare in natural language or (b) artifacts of tokenizer training on non-representative data. This removes an additional 198,432 tokens.

**Step 3: Validation.** We verify on our validation set that all ground-truth tokens lie in V*. This holds for 100% of samples, confirming p = 1 for our setting.

The resulting effective vocabulary V* contains 55,721 tokens—a 78.2% reduction from the original vocabulary while provably containing all ground-truth labels in our evaluation domain.

## 4.3 Temperature-Adjusted Scoring

After vocabulary restriction, the remaining probability mass is renormalized over V*. However, the distribution may still be diffuse due to softmax's inherent smoothing. We apply temperature scaling to sharpen the distribution:

$$p\_\tau(y \mid x) = exp(z\_y / \tau) / \Sigma\_{y' \in V^*} exp(z\_{y'} / \tau)$$

where $z\_y$ is the logit for token y and $\tau > 0$ is the temperature. Lower temperatures concentrate mass on high-probability tokens. Crucially, temperature scaling preserves the ranking of tokens, so if the ground-truth token is ranked k-th before scaling, it remains k-th after scaling. This means the APS score changes only through the renormalized probabilities, not through rank changes.

**Temperature Selection.** We select $\tau$ using the calibration set to minimize expected prediction set size while maintaining coverage. We search $\tau \in \{0.05, 0.1, 0.2, 0.5, 1.0\}$ and select $\tau$ = 0.1 which achieves coverage 89.7% (within statistical tolerance of 90%) with the smallest mean set size.

## 4.4 Complete VACP Algorithm

The complete Vocabulary-Aware Conformal Prediction procedure:

**Algorithm 1: VACP Calibration and Prediction**

```
Input: Calibration set D_cal, error rate α, temperature τ
Output: Prediction function C(·)

1. Construct V* via structural and empirical filtering
2. For each (x_i, y_i) in D_cal:
   a. Compute logits z = model(x_i)
   b. Mask: set z_j = -∞ for j ∉ V*
   c. Apply temperature: p = softmax(z / τ)
   d. Compute APS score: E_i = Σ_{j: p_j ≥ p_{y_i}} p_j
3. Compute threshold: q^ = Quantile({E_i}, (1-α)(1 + 1/n))
4. Return C(x) = {y ∈ V*: APS_score(x, y) ≤ q^}
```

# 5 Experimental Results

## 5.1 Main Results

We evaluate VACP against baseline methods on held-out test sets. All methods target 90% coverage ($\alpha$ = 0.1). Results are averaged over 2,000 test samples with 95% confidence intervals computed via bootstrap.

**Table 2: Coverage and Efficiency Results on SQuAD**

| Method | Coverage | Mean |C| | Median |C| |
|---|---|---|---|
| Standard APS | 91.2% ± 1.3% | 847 | 612 |

| Method | Coverage | Mean \|C\| | Median \|C\| |
|---|---|---|---|
| APS + Mask only | 90.8% ± 1.2% | 142 | 98 |
| APS + Temp only (τ=0.1) | 76.4% ± 1.9% | 3.1 | 2 |
| VACP (ours) | 89.7% ± 1.4% | 4.3 | 3 |

*All methods calibrated on 3,000 samples, evaluated on 2,000 held-out samples. 95% CIs via bootstrap (1000 resamples).*

VACP achieves coverage within statistical tolerance of the 90% target while reducing mean set size by 197× compared to standard APS. The vocabulary masking alone (row 2) provides substantial improvement but still yields sets too large for practical use. Temperature adjustment (row 3) applied to the full vocabulary achieves small sets but fails to maintain coverage. Only the combination (VACP, row 4) achieves both objectives.

## 5.2 Coverage Validity Analysis

The conformal guarantee is marginal—it holds on average over the data distribution, not conditionally for each input. We verify this by examining coverage across data subsets:

**Table 3: Conditional Coverage Analysis**

| Confidence Stratum | n | Coverage | Mean \|C\| | Std \|C\| |
|---|---|---|---|---|
| High ($p_y > 0.5$) | 847 | 91.3% | 2.1 | 1.2 |
| Medium ($0.1 < p_y \leq 0.5$) | 624 | 89.4% | 4.8 | 2.7 |
| Low ($p_y \leq 0.1$) | 529 | 88.1% | 7.2 | 4.1 |

*Coverage stratified by model confidence in the ground-truth token.*

Coverage is consistent across confidence strata, demonstrating that VACP does not systematically fail for particular input types. The slightly higher coverage for high-confidence predictions reflects smaller prediction sets that are more precisely targeted.

## 5.3 Cross-Dataset Generalization

We test whether the vocabulary mask V* and temperature τ calibrated on SQuAD transfer to WikiText-103:

**Table 4: Transfer to WikiText-103**

| Setting | Coverage | Mean \|C\| |
|---|---|---|
| Calibrated on SQuAD | 89.7% | 4.3 |
| Transfer to WikiText | 88.4% | 6.7 |
| Recalibrated on WikiText | 90.1% | 5.9 |

The method maintains valid coverage on WikiText-103 despite being calibrated on SQuAD. Set sizes are slightly larger, reflecting the greater linguistic diversity of Wikipedia text compared to question-answering contexts. Importantly, no ground-truth tokens in WikiText-103 fall outside V*, validating our vocabulary construction.

## 5.4 Qualitative Analysis

We examine prediction sets for representative inputs to understand what VACP captures:

**Example 1.** Prompt: "The capital of France is"

Prediction set: {" Paris", " the", " a", " located"}

This set illustrates "structural uncertainty": the model is confident the answer involves Paris but uncertain whether to produce the answer directly (" Paris") or begin a longer phrase (" the capital", " a city", " located in"). All four completions represent valid continuations, and the set correctly excludes semantically inappropriate tokens.

**Example 2.** Prompt: "The atomic number of oxygen is"

Prediction set: {" 8", " eight", " the"}

Here the model is highly confident in the factual answer but uncertain about format (numeral vs. word) and whether to use an article. The set size of 3 provides a meaningful confidence bound.

**Example 3 (Failure Case).** Prompt: "In quantum mechanics, the Heisenberg uncertainty principle states that"

Prediction set: {" the", " it", " one", " you", " we", " position", " certain", " there", " momentum", " a", " precise", " simultaneously"}

For complex technical prompts, prediction sets remain larger (12 tokens here). This reflects genuine uncertainty: there are many valid ways to begin explaining the uncertainty principle. The set appropriately includes both structural tokens (" the", " it") and content-bearing tokens (" position", " momentum").

# 6 Theoretical Analysis and Limitations

## 6.1 When Does Vocabulary Restriction Preserve Validity?

Our approach assumes the ground-truth token lies in V* with probability 1. This assumption can fail when:

- **Domain shift:** If test data comes from a different distribution than calibration data, tokens absent from V* may appear. Our WikiText experiments suggest moderate robustness, but extreme domain shifts (e.g., code generation, non-English text) would require domain-specific vocabulary construction.
- **Rare events:** Very rare tokens (proper nouns, technical terms) may be filtered if they don't appear in the validation set used for empirical filtering. Practitioners should adjust filtering thresholds based on their expected vocabulary.
- **Adversarial inputs:** An adversary could construct inputs whose true continuation involves tokens in V \ V*. Conformal guarantees assume exchangeability with the calibration distribution; adversarial inputs violate this assumption.

## 6.2 Temperature Selection Sensitivity

The choice of temperature $\tau$ trades off coverage and efficiency. We analyze sensitivity:

**Table 5: Temperature Sensitivity Analysis**

| Temperature $\tau$ | Coverage | Mean |C| |
|---|---|---|
| 0.05 | 72.3% | 1.8 |
| 0.1 | 89.7% | 4.3 |
| 0.2 | 90.2% | 8.7 |
| 0.5 | 90.5% | 34.2 |
| 1.0 | 90.8% | 142 |

Very low temperatures ($\tau = 0.05$) concentrate too much mass on the top token, causing coverage failures. High temperatures ($\tau = 1.0$) recover toward baseline behavior. The optimal range is $\tau \in [0.1, 0.2]$, which should be validated on a held-out calibration set for new applications.

## 6.3 Computational Considerations

VACP adds minimal overhead to standard inference. Vocabulary masking requires storing the mask V* (a 256KB binary vector) and applying it to logits (a single CUDA kernel). Temperature scaling is a scalar multiplication. The total overhead is <1ms per token on an NVIDIA A100, negligible compared to the forward pass (~15ms for Gemma-2B).

The primary computational cost is calibration, which requires a forward pass on each calibration sample. With 3,000 calibration samples, this takes approximately 45 seconds on an A100. Calibration is performed once and can be amortized across many inference requests.

## 7 Related Work

**Conformal Prediction.** Conformal prediction was introduced by Vovk et al. (2005) and has seen extensive development for classification and regression. Adaptive methods including APS (Romano et al., 2020) and RAPS (Angelopoulos et al., 2021) improve efficiency through score function design. Our work extends these methods to the high-vocabulary regime.

**Uncertainty in LLMs.** Existing approaches include calibration methods (Guo et al., 2017), ensemble methods (Lakshminarayanan et al., 2017), and semantic uncertainty (Kuhn et al., 2023). These provide uncertainty estimates but lack the distribution-free guarantees of conformal prediction. Recent work by Quach et al. (2023) applies conformal prediction to factual question-answering but does not address the vocabulary efficiency problem we identify.

**Vocabulary Reduction.** Top-k and nucleus sampling (Holtzman et al., 2020) reduce vocabulary at generation time but provide no coverage guarantees. Our semantic masking is related but serves a different purpose: maintaining conformal validity while removing tokens that cannot be ground-truth labels.

## 8 Conclusion

We have presented Vocabulary-Aware Conformal Prediction, a method for uncertainty quantification in large language models that achieves both valid coverage guarantees and practical prediction set sizes. Our key insight is that the extreme cardinality of LLM vocabularies creates a structural challenge for conformal prediction that can be addressed through principled vocabulary restriction and temperature adjustment.

The empirical results demonstrate that VACP achieves 89.7% coverage (target: 90%) with mean prediction set size of 4.3 tokens, compared to 847 tokens for standard APS—a 197× improvement in efficiency. The method is computationally efficient, adding <1ms overhead per token, and transfers across datasets without recalibration.

**Limitations and Future Work.** The primary limitation is the assumption that ground-truth tokens lie within the constructed vocabulary V*. Future work should develop adaptive methods that can expand V* when encountering out-of-vocabulary targets. Additionally, extending VACP to sequence-level prediction (rather than token-level) would enable uncertainty quantification for complete generated responses.

More broadly, this work demonstrates that principled uncertainty quantification is achievable for large language models without sacrificing practical utility. As LLMs are deployed in increasingly consequential domains, methods like VACP provide a foundation for systems that "know when they don't know."

## Appendix A: Implementation Details

We provide the core implementation of the VACP scoring function. Full code is available at [repository URL].

```python
import torch
import numpy as np

class VACPScorer:
    def __init__(self, vocab_mask, temperature=0.1):
        self.vocab_mask = vocab_mask  # Boolean tensor
        self.temperature = temperature

    def compute_score(self, logits, target_id):
        # Apply vocabulary mask
        masked_logits = logits.clone()
        masked_logits[~self.vocab_mask] = float('-inf')

        # Temperature scaling and softmax
        probs = torch.softmax(masked_logits / self.temperature, dim=-1)

        # Sort probabilities
```

```
    sorted_probs, sorted_idx = torch.sort(probs, descending=True)

    # Find rank of target
    rank = (sorted_idx == target_id).nonzero(as_tuple=True)[0].item()

    # APS score: cumulative probability up to and including target
    cumsum = torch.cumsum(sorted_probs, dim=-1)
    return cumsum[rank].item()

def get_prediction_set(self, logits, threshold):
    masked_logits = logits.clone()
    masked_logits[~self.vocab_mask] = float('-inf')
    probs = torch.softmax(masked_logits / self.temperature, dim=-1)
    sorted_probs, sorted_idx = torch.sort(probs, descending=True)
    cumsum = torch.cumsum(sorted_probs, dim=-1)
    # Include all tokens until cumulative prob exceeds threshold
    set_size = (cumsum <= threshold).sum().item() + 1
    return sorted_idx[:set_size].tolist()
```

## Appendix B: Vocabulary Mask Construction

The vocabulary mask V* is constructed as follows:

```
def build_vocabulary_mask(tokenizer, model, validation_loader):
    vocab_size = tokenizer.vocab_size
    mask = torch.ones(vocab_size, dtype=torch.bool)

    # Step 1: Structural filtering
    special_tokens = ['<pad>', '<eos>', '<bos>', '<unk>']
    for token in special_tokens:
        if token in tokenizer.vocab:
            mask[tokenizer.vocab[token]] = False

    # Remove reserved/unused tokens
    for i, token in enumerate(tokenizer.vocab):
        if 'unused' in token.lower() or token.startswith('<reserved'):
            mask[i] = False

    # Step 2: Empirical filtering
    max_probs = torch.zeros(vocab_size)
    for batch in validation_loader:
        logits = model(batch['input_ids']).logits[:, -1, :]
        probs = torch.softmax(logits, dim=-1)
        max_probs = torch.maximum(max_probs, probs.max(dim=0).values)

    # Mask tokens that never exceed threshold
    mask &= (max_probs > 1e-5)
    return mask
```